# Extended Two-Dimensional PCA for Efficient Face Representation and Recognition

M. Safayani, M.T. Manzuri Shalmani, M. Khademi
*Computer Engineering Department of Sharif University of technology*
*safayani@ce.sharif.edu, manzuri@sharif.edu, khademi@ce.sharif.edu*

**Abstract**

*In this paper a novel method called Extended Two-Dimensional PCA (E2DPCA) is proposed which is an extension to the original 2DPCA. We state that the covariance matrix of 2DPCA is equivalent to the average of the main diagonal of the covariance matrix of PCA. This implies that 2DPCA eliminates some covariance information that can be useful for recognition. E2DPCA instead of just using the main diagonal considers a radius of r diagonals around it and expands the averaging so as to include the covariance information within those diagonals. The parameter r unifies PCA and 2DPCA. r=1 produces the covariance of 2DPCA, r=n that of PCA. Hence, by controlling r it is possible to control the trade-offs between recognition accuracy and energy compression (fewer coefficients), and between training and recognition complexity. Experiments on ORL face database show improvement in both recognition accuracy and recognition time over the original 2DPCA.*

## 1. Introduction

One of the most important techniques in image representation and feature extraction is Principal Component Analysis (PCA)[1]. The first step of PCA is to transform the matrix associated to an image into vectors of high dimension and then compute the covariance matrix in high-dimensional vector space. The Covariance matrix can not be evaluated accurately due to its large size and the relatively small number of training samples [2]. Furthermore, computing the correspondence eigenvectors is very time consuming. For solving these problems recently a new technique called two-dimensional PCA (2DPCA) has been proposed [2]. The main idea of this method is that it eliminates the image to vector conversion stage and computes covariance matrix directly based on the 2D images. Because of small size of the covariance matrix and adequate training samples, 2DPCA can compute the eigenvectors faster and much more accurate than PCA [2]. Recently in [3] has been shown that 2DPCA is equivalent to block based approach which each block is one line of the image. Although Yang in [2] showed the better performance of matrix-based representation than traditional one-dimensional vector-based approach, there are also some drawbacks in this method. First, it needs to more coefficients for representing an image than previous one-dimensional methods [2]. Kong in [4] tries to solve this problem by proposing two sets of projection directions called Generalized two-dimensional PCA. For finding these projection directions he proposed an iterative algorithm theoretically having local optimal solutions and was dependent on the initial conditions. Second, it loses the covariance information between different local geometric structures in the image while PCA preserves such information which is important for recognition [5]. In this paper we propose a novel method called extended two-dimensional PCA (E2DPCA) which is more useful and efficient for real application. We expose the mathematical relationship between the covariance matrix of the PCA and that of 2DPCA and try to evaluate a new covariance matrix preserving more local structure information than previous 2DPCA. This new covariance matrix can be evaluated more accurately than both that of one and two-dimensional approaches. Also, E2DPCA represents an image with much fewer coefficients which result in more computational efficiency than 2DPCA.

The remaining parts of the paper are organized as follows. In section 2 PCA and 2DPCA and their mathematical relationship is described. The proposed E2DPCA is introduced in section 3. In section 4 experimental results are presented. Finally, we have a conclusion in section 5.

## 2. 2DPCA and its relationship to PCA

### 2.1. PCA

In PCA image A, a $m \times n$ random matrix, first is reshaped into a high-dimensional vector, $a \in (m*n) \times 1$, by concatenating its columns or its rows, then the scatter matrices are computed as follows:

$$S^{1D} = \frac{1}{M} \sum_{j=1}^{M} [a_j - \bar{a}][a_j - \bar{a}]^T$$

Where there is $M$ training image samples in the database. $a_j \in (m*n) \times 1$ is the $jth$ reshaped image, and $\bar{a}$ denotes the mean of whole training set. The optimal projection axis is the set of eigenvectors of $S^{1D}$ corresponding to the first $d$ largest eigenvalues. For more details see [1].

### 2.2. 2DPCA

2DPCA projects the image matrix A, onto $X$ by $Y = AX$ linear transformation. The total scatter of the projected samples, $S_x$, can be used for determining the optimal projection.

$$S_x = x^T E\{[A - EA][A - EA]^T\} x = x^T S^{2D} x$$

Where $S^{2D} = E\{[A - EA][A - EA]^T\}$, called the image covariance matrix can directly be evaluated by using M training samples. Let the average image of all training samples be denoted by $\bar{A}$, then $S^{2D}$ can be evaluated by:

$$S^{2D} = \frac{1}{M} \sum_{j=1}^{M} [A_j - \bar{A}][A_j - \bar{A}]^T$$

The optimal projection axis $X_{Opt}$ is composed by the orthonormal eigenvector of the $S^{2D}$ corresponding to the $d$ largest eigenvalues.

### 2.3. Mathematical relationships between PCA and 2DPCA

As stated in [3], 2DPCA can work in either the row or column direction of the images. Here without loss of generality we continue based on the column direction of the images. Suppose $A_j$, a $m \times n$ random image matrix, is the $jth$ sample in the database, and $\bar{A} = (1/M) \sum_{j=1}^{M} A_j$ is the global mean matrix. Let

$$A_j = [A_j(1), A_j(2), ..., A_j(n)]$$

$$\bar{A} = [\bar{A}(1), \bar{A}(2), ..., \bar{A}(n)]$$

Where $A_j(i)$ and $\bar{A}(i)$ are the $ith$ column vector of $A_j$ and $\bar{A}$, respectively. It's easy to show that [3]:

$$S^{2D} = \frac{1}{M} \sum_{j=1}^{M} \sum_{i=1}^{n} (A_j(i) - \bar{A}(i))(A_j(i) - \bar{A}(i))^T \quad (1)$$
$$= S_{1,1}^{2D} + S_{2,2}^{2D} + ... + S_{n,n}^{2D}$$

where $S_{i,p}^{2D}$ is defined as:

$$S_{i,p}^{2D} = \frac{1}{M} \sum_{j=1}^{M} (A_j(i) - \bar{A}(i))(A_j(p) - \bar{A}(p))^T$$

$$i, p = 1, ..., n$$

$S_{i,i}^{2D} \in R^{m \times m}$ is the scatter matrix of the $ith$ column of the training images in the database.

**Theorem 1**, $S^{1D} \in R^{(m*n) \times (m*n)}$ is partitioned into submatrices as indicated below:

$$S^{1D} = \begin{bmatrix} S_{1,1}^{2D} & S_{1,2}^{2D} & \cdots & S_{1,n}^{2D} \\ S_{2,1}^{2D} & S_{2,2}^{2D} & \cdots & S_{2,n}^{2D} \\ \vdots & \vdots & \vdots & \vdots \\ S_{n,1}^{2D} & S_{n,2}^{2D} & \cdots & S_{n,n}^{2D} \end{bmatrix} \quad (2)$$

**Proof:** Let

$$a_j = [A_j(1)^T, A_j(2)^T, ..., A_j(n)^T]^T$$
$$\bar{a} = [\bar{A}(1)^T, \bar{A}(2)^T, ..., \bar{A}(n)^T]^T$$
$$B_j(i) = (A_j(i) - \bar{A}(i))$$

Therefore we can express $S^{1D}$ as

$$S^{1D} = \frac{1}{M} \sum_{j=1}^{M} (a_j - \bar{a})(a_j - \bar{a})^T$$

$$= \frac{1}{M} \sum_{j=1}^{M} \begin{bmatrix} B_j(1) \\ \vdots \\ B_j(n) \end{bmatrix} [B_j(1)^T, ..., B_j(n)^T]$$

$$= \frac{1}{M} \begin{bmatrix} \sum_{k=1}^{M} B_k(1) B_k(1)^T & \cdots & \sum_{k=1}^{M} B_k(1) B_k(n)^T \\ \vdots & \vdots & \vdots \\ \sum_{k=1}^{M} B_k(n) B_k(1)^T & \cdots & \sum_{k=1}^{M} B_k(n) B_k(n)^T \end{bmatrix}$$

By substituting $S_{i,p}^{2D} = \frac{1}{M} \sum_{j=1}^{M} B_j(i) B_j(p)^T$ into above equation, theorem is proved.

## 3. Extended Two-Dimensional PCA (E2DPCA)

We can see from equations (1) that $S^{2D}$ is modeled by sum of all scatter matrices of different column indices in the main diagonal of $S^{1D}$ in term of equality (2). It can be seen that in $S^{2D}$ many important information between different local structures of the images has been eliminated which may have important discriminative information for recognition. On the other hand, retaining all this information leads to a large covariance matrix needing to a large number of training data for correctly evaluation. In our method instead of just using the main diagonal, we consider a radius of r diagonals around it and expand the averaging so as to include the covariance information within those diagonals. Moreover, the parameter r unifies PCA and 2DPCA. r=1 produces the covariance of 2DPCA, r=n that of PCA. Hence, by controlling r it is possible to control the trade-offs between recognition accuracy and energy compression (fewer coefficients), and between training and recognition complexity: PCA has lower recognition rates but produces fewer coefficients, and it is harder to compute the PCA covariance but the projection in the resulting eigenvectors is easier. The new covariance matrix when r=2 :

$$S^{E2D} = \sum_{i=1}^{n/2} \begin{bmatrix} S^{2D}_{2i-1,2i-1} & S^{2D}_{2i-1,2i} \\ S^{2D}_{2i,2i-1} & S^{2D}_{2i,2i} \end{bmatrix} \quad (3)$$

$S^{E2D} \in R^{2m \times 2m}$ is a symmetric matrix. It is four times larger than that of 2DPCA and preserves more local structure information. More precisely, new covariance matrix not only keep covariance information from the main diagonal of $S^{1D}$ in term of equality (2), but also keep some information from one diagonal above and below it. Although the new covariance matrix becomes larger, extracting eigenvectors from it is still easy task. We should mention that parameter r has a natural interpretation. It is the number of image columns that are stacked for the computation of the PCA transform. In PCA (r=n), the n columns are stacked into a column vector. In 2DPCA (r=1) there is no stacking and the image is represented as a matrix. When r=k, k columns are stacked at a time. Figure 1, Shows an example when r=2. in this situation, we stack two columns at a time, making the matrix twice as tall and half as fat. When the number of columns or rows is odd we add zero lines to make it even. We maximize $J^{E2D}(w) = w^T S^{E2D} w$ for finding the optimal projection direction. It's well known that the optimal direction is equal to the eigenvectors of $S^{E2D}$. These eigenvectors are used to form a $(2*m) \times d$ matrix $D = [X_1,...,X_d]$. These optimal projection vectors are used for feature extraction. For a given test face image $A$, we at first reshape it in the way as illustrated in figure (1.a) to form an image matrix $B$, and then use the following equation:

$$Y_k = B^T X_k \qquad k = 1,2,...,d$$

Dimension of B is $(2*m) \times (n/2)$ so the dimension of $Y_k$ becomes $(n/2) \times 1$, by selecting d vector we have a feature matrix of $(n/2) \times d$ which is fewer than feature matrix of 2DPCA $(n \times d)$. A nearest neighbor classifier with Euclidean distance is used for classification. It's obvious that when r is greater than 2, the original $m \times n$ image matrices transform into $(r*m) \times (n/r)$ matrix and the resulted covariance matrix have $(r*m) \times (r*m)$ dimension and after feature extraction the dimension of feature matrix is $(n/r) \times d$ which is much less than the dimension of 2DPCA feature matrix $(n \times d)$. By increasing r, covariance matrix becomes larger, and feature matrix becomes smaller which result in the decreasing of computation cost. In PCA, finding optimal projection is time consuming but recognition is very fast. Reverse is true for 2DPCA. In E2DPCA we have a trade off between the computation cost in eigenvector decomposition and recognition.

## 4. Experiments

For showing the effect of E2DPCA, we use ORL database [6]. This database contains images from 40 individuals, each providing 10 different images. The pose, expression, and facial details variations are also included. The images are taken with a tolerance for some tilting and rotation of the face of up to 20 degrees. Moreover, there is also some variation in the scale of up to about 10 percent. All images are grayscale and normalized to a resolution of $112 \times 92$ pixels. In our experiments we use first five image sample per class for training, and the remaining images for the test. Thus the total number of training samples and testing samples were both 200. 2DPCA which originally works on row direction of images [4] can also work on the column direction. In our experiments, we use the word "alternative" for expressing column based algorithm. When we want to apply E2DPCA in row direction of the image we should transforms the image in the way as illustrated in figure (1.b). For finding the optimal point E2DPCA is applied with different $r$. Table (1) shows the comparisons of six methods PCA, 2DPCA (row based), alternative 2DPCA (column based), E2DPCA (row based), alternative E2DPCA (column based) and 2D2PCA [7] on recognition accuracy and dimension of feature vectors. As can be seen from this table top accuracy of E2DPCA is better than that of other methods which is because of using more local geometric structure information. Moreover, it represents image with fewer coefficients than both 2DPCA and 2D2PCA. Table (2) compares dimensions of the feature vectors and recognition time cost of the different method under the same accuracy. It can be seen from this table that E2DPCA represents the image with much fewer coefficients than both 2DPCA and 2D2PCA, for instance 2DPCA, 2D2PCA and E2DPCA reach same accuracy with 896, 144 and 50 coefficients, respectively. Also recognition time in E2DPCA is less than other methods.

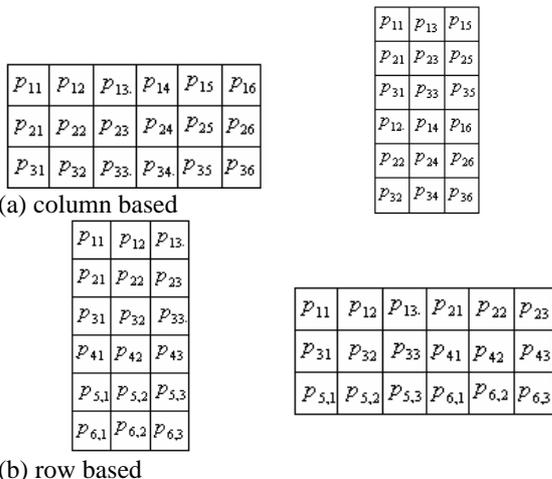

(a) column based

(b) row based

Figure 1. Illustration of the way for deriving the reshaped image for E2DPCA, Left: original image, Right: reshaped image

Table 1. Comparison of top recognition accuracy in different methods on ORL database

| Method | (%) | Dimension |
|---|---|---|
| PCA ( Eigenface) | 85.0 | 34 |
| 2DPCA | 91.5 | 112*8=896 |
| Alternative 2DPCA | 91.5 | 92*4=368 |
| 2D2PCA | 92.0 | 16*16=256 |
| E2DPCA (r=21) | 93.0 | 6*20=120 |
| Alternative E2DPCA (r=6) | 92.5 | 16*18=288 |

Table 2. Comparison of dimensionality of different methods under the same accuracy on ORL database

| Method | (%) | Dimension | Time (s) |
|---|---|---|---|
| 2DPCA | 91.5 | 112*8=896 | 2.37 |
| Alternative 2DPCA | 91.5 | 92*4=368 | 1.74 |
| 2D2PCA | 91.5 | 12*12=144 | 1.70 |
| E2DPCA (r=23) | 91.5 | 5*10=50 | 1.48 |
| Alternative E2DPCA (r=23) | 91.5 | 4*16=64 | 1.65 |

## 5. Conclusion

In this paper a novel method called E2DPCA was proposed which is an extension to the current 2DPCA algorithm. We try to unify PCA and 2DPCA by defining a parameter notated as r which it extends the covariance matrix of 2DPCA toward that of PCA. 2DPCA eliminates much local geometric structure information presented in the covariance matrix of PCA. This information is important for recognition so we defined an adjacent radius which let us to adjust how much this information being used. Experimental results show improvement in both recognition accuracy and recognition time over 2DPCA.